\documentclass{article} 
\usepackage{nips12submit_e,times}

\usepackage{algorithm}
\usepackage{algorithmic}
\usepackage{graphicx}

\title{Temporal Autoencoding Restricted Boltzmann Machine}

\author{ Chris H\"ausler\\ Neuroinformatics and Theoretical Neuroscience Group\\ Freie
Universit\"at Berlin, Bernstein Center for Computational Neuroscience\\ Berlin, Germany \\
\texttt{chausler@gmail.com} \\ \And Alex Susemihl \\
Department of Artificial Intelligence \\ Berlin Institute of
Technology, Bernstein Center for Computational Neuroscience\\ Berlin, Germany\\
\texttt{alexsusemihl@gmail.com} \\ }

\nipsfinalcopy 

\begin{document}

\maketitle

\begin{abstract} Much work has been done refining and
characterizing the receptive fields learned by deep
learning algorithms. A lot of this work has focused on the
development of Gabor-like filters learned when enforcing
sparsity constraints on a natural image dataset. Little work
however has investigated how these filters might expand to the
temporal domain, namely through training on natural movies.
Here we investigate exactly this problem in established
temporal deep learning algorithms as well as a new learning
paradigm suggested here, the Temporal Autoencoding Restricted Boltzmann Machine (TARBM).  \end{abstract}

\section{Introduction}

In the early days of Machine Learning, feature extraction was usually approached in a task-specific way.
The complexity and high dimensionality involved in doing so in an unsupervised fashion was seen
as a major barrier and expert features were thought to yield the best results for classification and
representation tasks \cite{fukunaga1990introduction}.
Recently however, a number of advances have brought the field of unsupervised feature extraction back
into the center stage of machine learning. Increases in computational power, allowing for algorithms
trained on very large datasets, together with new techniques to train deep architectures have
yielded insightful results in unsupervised feature learning even in uncurated sets of natural images
\cite{le2012}. Examples of such algorithms are denoising Autoencoders (dAEs) and Restricted Boltzmann Machines (RBMs) \cite{le2012,mohamed2011deep,hinton2012improving}.\par

In unsupervised feature learning, it is the structure of the data that defines the features to be learnt by a given model. In Computational Neuroscience, this link between the ensemble of natural stimuli an organism is exposed to and the shape of the tuning functions in their sensory systems has been a subject of great interest \cite{karklin2011, lewicki2002, sprekeler2011}. Specifically in the field of vision neuroscience, a number of principles have been proposed to explain the shape of tuning functions in primary visual cortex based on the properties of natural images, for example redundancy minimization \cite{atick1992} and predictive coding \cite{rao1999}. In recent years, it has been shown that simple unsupervised learning algorithms such as Sparse Coding, dAEs and RBMs can also be used to learn structure from natural stimuli, independently of labels and supervision, and that the types of structure learnt can be related back to cortical receptive fields found in the mammalian brain \cite{saxe2011, lee2008} .
\par

While most of this research in vision has focused on finding optimal filters for representing and decoding sets of static natural images \cite{olshausen1996,bell1997}, here we seek to understand how these optimal filters extend to the temporal domain. We build on existing work in the field to develop the Temporal Autoencoding Restricted Boltzmann Machine (TARBM) and show that it is able to learn high level structure in a natural movie dataset and account for the transformation of these features over time.

\par



\section{Existing Models}
Restricted Boltzmann Machines (RBMs) \cite{hinton2006fast, hinton2006reducing} and Autoencoders (AEs) \cite{bengio2007, ranzato2007} have in recent years become prominent methods of unsupervised feature learning with applications in a wide variety of machine learning fields.
As both of these models are well known and discussed at length in many other papers, we will introduce them only briefly here.
\par
Both models are two-layer neural networks, all to all connected between the layers but with no intralayer connectivity. The models consist of a visible and a hidden layer, where the visible layer represents the input to the model whilst the hidden layer's job is to learn a meaningful representation of the data in some other dimensionality. We will represent the visible layer activation variables by $v_i$, the hidden activations by $h_j$ and the vector variables by $\mathbf{ v} = \{v_i\}$ and $\mathbf{h} = \{h_j\}$. 

\begin{figure} \begin{center}
\includegraphics[width=0.45\columnwidth]{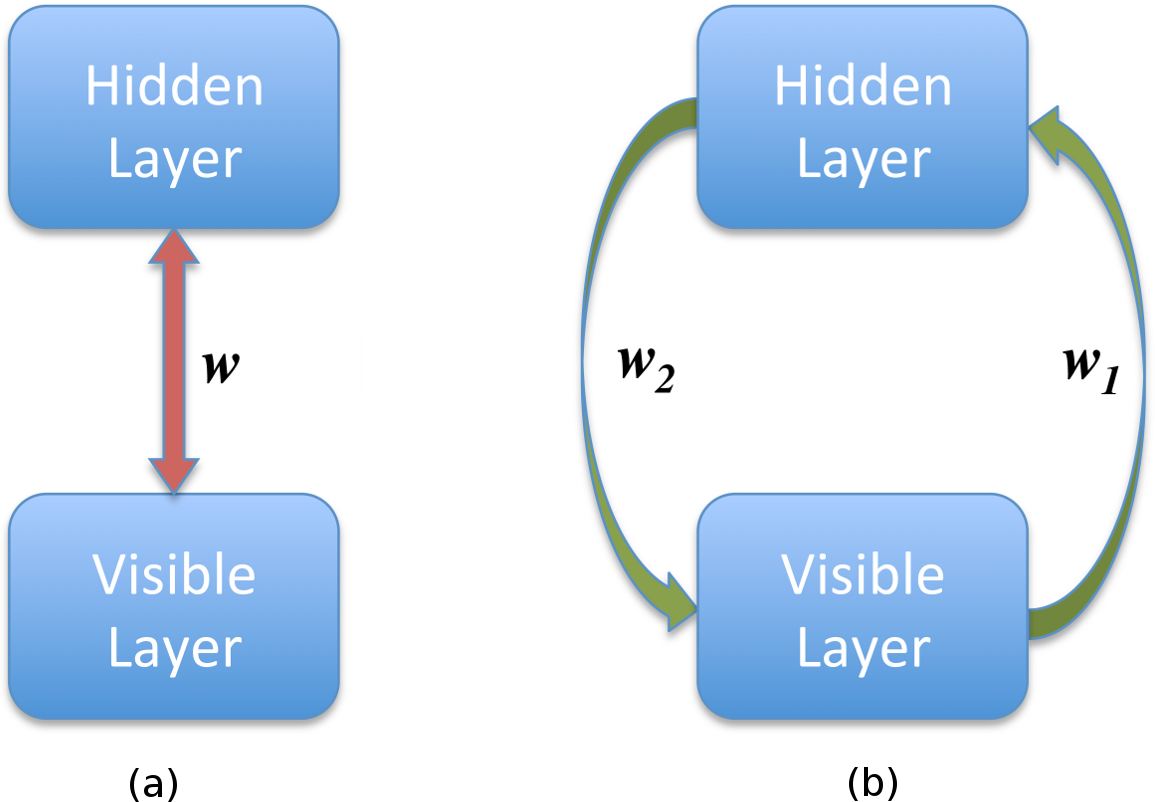}
\caption{Restricted Boltzmann Machine (a) and Autoencoder (b) architectures}
\label{fig:rbm_and_autoencoder} \end{center}
\end{figure}

Autoencoders are a deterministic model with two weight matrices $\mathbf{w}_{1}$ and $\mathbf{w}_{2}$ representing the flow of data from the visible-to-hidden and hidden-to-visible layers respectively (see figure \ref{fig:rbm_and_autoencoder}b). AEs are trained to perform optimal reconstruction of the visible layer, often by minimizing the mean-squared error (MSE) in a reconstruction task. This is usually evaluated as follows: Given an activation pattern in the visible layer $\mathbf{v}$, we evaluate the activation of the hidden layer by $\mathbf{h} = sigm(\mathbf{v}^\top \mathbf{w}_{1} + \mathbf{b}_{h})$. These activations are then propagated back to the visible layer through $\mathbf{\hat{v}} = sigm(\mathbf{h} \mathbf{w}_{2}^\top + \mathbf{b}_{v})$ and the weights $\mathbf{w}_{1}$ and $\mathbf{w}_{2}$ are trained to minimize the distance measure between the original and reconstructed visible layers. For example, using the squared euclidian distance we have a cost function of
$$
\mathcal{L}(\mathbf{w}_1,\mathbf{w}_2,\mathbf{b}^v,\mathbf{b}^h, \{\mathbf{v}^d\}) = \sum_{d} \| \mathbf{v}^d - \hat{\mathbf{v}}^d\|^2,
$$
where we have denoted the dataset by $\{\mathbf{ v}^d\}$ and the biases of the visible and hidden layer as $\mathbf{b}_{v}$ and $\mathbf{b}_{h}$ respectively. The weights can then be learned through stochastic gradient descent on the cost function.\par

Restricted Boltzmann Machines on the other hand are a stochastic model that assumes symmetric connectivity between the visible and hidden layers (see Figure \ref{fig:rbm_and_autoencoder}a) and seeks to model the structure of a given dataset. They are generally viewed as energy-based models, where the energy of a given configuration of activations $\{v_i\}$ and $\{h_j\}$ is given by
$$
E_{RBM}(\mathbf{v},\mathbf{h}|\mathbf{w},\mathbf{b}_v,\mathbf{b}_h) = - \mathbf{v}^\top \mathbf{w} \mathbf{h} - \mathbf{b}_v^\top \mathbf{v} - \mathbf{b}_h^\top \mathbf{h}.
$$
RBMs are usually trained through contrastive divergence, the central idea of which is to stabilize the transient induced by the presentation of data to the visible layer, therefore representing it in the hidden layer optimally. In practice this is achieved by learning the weights via the difference between the \emph{transient} and the \emph{equilibrium} correlations between visible and hidden layers. Sample correlations in the first presentation are taken as a proxy for the transient and correlations after $n$ successive Gibbs samples are taken as a proxy for the equilibrium correlation. The weight update is then defined as
$$
\Delta w_{i,j} \propto \left<v_i h_j\right>_{0}-\left<v_i h_j\right>_{n}.
$$
A number of auxiliary strategies have been used to improve the training process of RBMs such as mini-batch training, free energy minimization, Parzen windows, early stopping and sparsity constraints. In addition, RBMs can be stacked to form what is called a Deep Belief Network (DBN) \cite{hinton2006reducing} where each additional RBM models the output of the previous one to form a more abstract/high level representation.
\par
To date, a number of RBM based models have been proposed to capture the sequential structure in time series data. Two of these models, the Temporal Restricted Boltzmann Machine and the Conditional Restricted Boltzmann machine, are introduced below.

\subsection{Temporal Restricted Boltzmann Machine (TRBM)}
\begin{figure} \begin{center}
\includegraphics[width=\columnwidth]{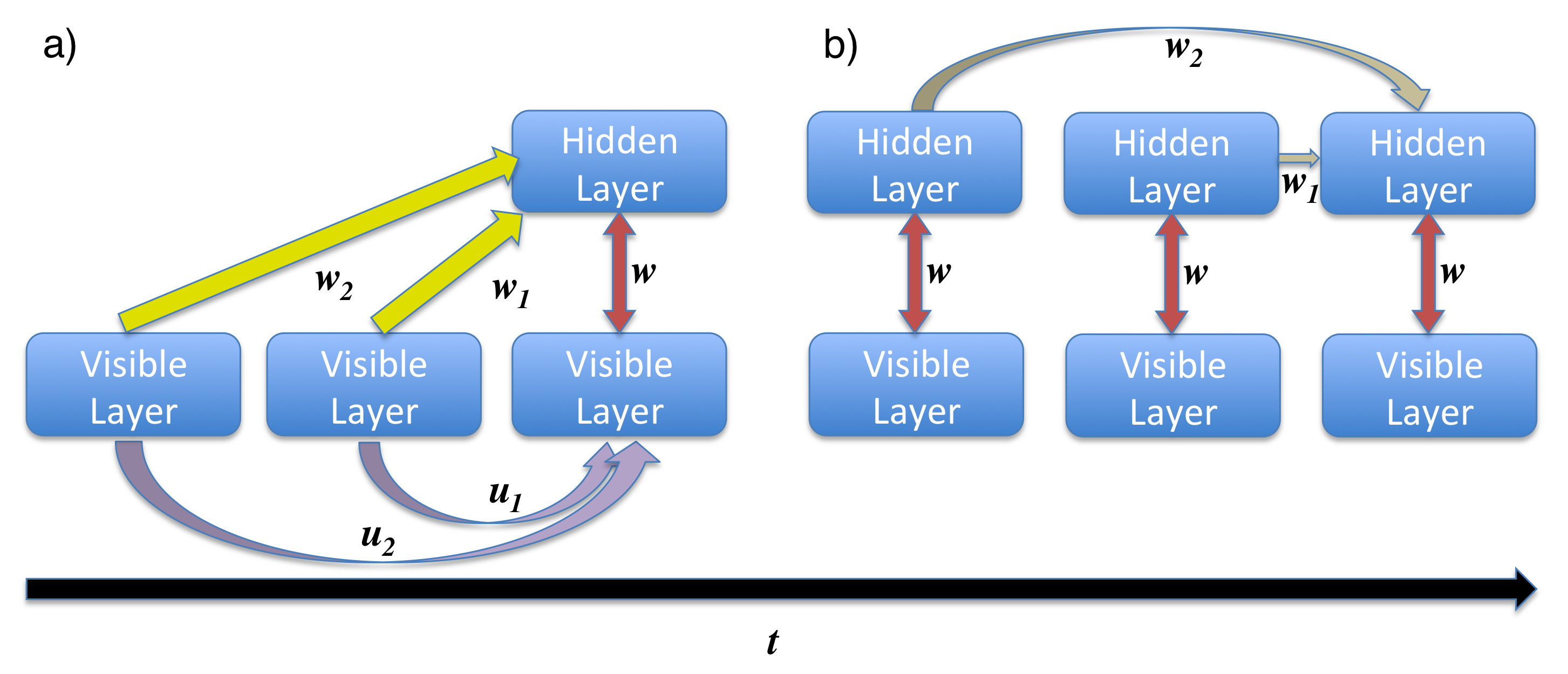}
\caption{(a) Conditional Restricted Boltzmann Machine Architecture (b) Architecture used by the Temporal RBM and the Temporal Autoencoding RBM } \label{fig:tarbm}
\end{center} \end{figure}

The Temporal Restricted Boltzmann Machine \cite{sutskever2007learning} is a temporal extension of the standard RBM whereby feed forward connections are included from previous time steps between hidden layers, from visible-to-hidden layers and from visible-to-visible layers. Learning is conducted in the same manner as a normal RBM using contrastive divergence and it has been shown that such a model can be used to learn non-linear system evolutions such as the dynamics of a ball bouncing in a box \cite{sutskever2007learning}. A more restricted version of this model, discussed in \cite{sutskever2008recurrent} can be seen in Figure \ref{fig:tarbm}b and only contains temporal connections between the hidden layers.
\par
If we denote by $\mathbf{h} = \{\mathbf{h}_0,\mathbf{h}_1,\ldots, \mathbf{h}_M\}$ the hidden layers and by $\mathbf{v} = \{\mathbf{v}_0,\mathbf{v}_1,\ldots,\mathbf{v}_M\}$ the visible layers,the energy of the model is given by
\begin{equation}
E(\mathbf{h,v|W}) = \sum_{i=0}^M E_{RBM}(\mathbf{ h}^{i},\mathbf{v}^{i}|\mathbf{w,b}) - \sum_{i=1}^M (\mathbf{h}^0)^\top \mathbf{w}_{i} \mathbf{h}^i,
\label{eqn:energy_trbm}
\end{equation}
where the weights are as given in Figure \ref{fig:tarbm}b. We denoted $\mathbf{W} = \{\mathbf{w},\mathbf{w}_1, \ldots, \mathbf{w}_{M}\}$, where $\mathbf{w}$ are the static weights and $\mathbf{w}_{1}$ to $\mathbf{w}_{M}$ are the delayed weights. These models have been shown to be amenable to stacking in deep architectures in the same manner as RBMs and AEs.

\subsection{Conditional Restricted Boltzmann Machine (CRBM)}
The Conditional Restricted Boltzmann Machine described in \cite{taylor2007modeling} contains no temporal connections from the hidden layer but includes connections from the visible layer at previous time steps to the current hidden and visible layers. The model architecture can be seen in Figure \ref{fig:tarbm}a. Again, learning with this architecture requires only a small change to the energy function of the RBM and can be achieved through contrastive divergence. The CRBM is likely the most successfull of the temporal RBM models to date and has been shown to both model and generate data from complex dynamical systems such as human motion capture data and video textures \cite{taylor2009composable}.

\section{Temporal Autoencoding Restricted Boltzmann Machines (TARBM)}

Here we present a new model, the TARBM, an extension of the Temporal RBM (with only hidden-to-hidden temporal connections) where a denoising Autoencoder approach is used to pretrain the temporal weights. We show that this approach provides a marked advantage over contrastive divergence training alone and that our model is able to outperform both the TRBM and CRBM on a classical temporal sequence task while yielding a deeper insight into the temporal representation of natural image sequence data.


\subsection{The Model}

Much of the motivation for this work is to gain insight into the typical evolution of learned hidden layer features present in natural movie stimuli. With the CRBM this is not possible as it is unable to explicitly model the evolution of hidden features without resorting to a deep network architecture.
We address this by  using a layerwise approach, much in the same vein as that used when stacking RBMs to form a Deep Belief Network \cite{hinton2006reducing}, but through time. We stack a given number of RBMs side by side in time and train the temporal connections between the hidden layers  (see Figure \ref{fig:tarbm}b) to minimize the reconstruction error, in a process similar to Autoencoder training \cite{bengio2007}. A simple autoregressive model is used to account for the dynamics of the hidden layer allowing us to train a \emph{dynamic prior} over the temporal evolution of the stimulus.
\par


\subsection{Training Algorithm}

We model our network as an energy-based function with interactions between the hidden layers at different time lags. The energy of the model is given by Equation \ref{eqn:energy_trbm} as in the case of the TRBM and is essentially an $M$-th order autoregressive RBM model and can be trained through standard contrastive divergence. The individual RBM visible-to-hidden weights $\mathbf{w}$ are initialized through contrastive divergence with a sparsity constraint on static samples of the dataset. After that, to ensure that the weights representing the hidden-to-hidden connections ($\mathbf{w}_{d}$) encode the dynamic structure of the ensemble, we initialize them by pre-training in the fashion of a denoising Autoencoder. We do this by treating the visible layer activation at time $t-d$, where $d$ is the temporal delay, as a corrupted version of the \textit{true} visible activation at time $t$. With this view, the model should learn to reconstruct the visible layer at time $t$ by transforming the \textit{corrupted} input at $t-d$ through the model as in the case of a denoising Autoencoder. The pretraining is described in Algorithm \ref{alg:pretraining}.

\begin{algorithm}[htb]
\caption{Pre-Training Temporal weights through Autoencoding}
\label{alg:pretraining}
\begin{algorithmic}
\FOR{each sequence of images $I(t-d),\ldots,I(t)$, we take $\mathbf{v}^0 = I(t), \ldots, \mathbf{v}^d = I(t-d)$ and}
\FOR{$d = 1$ \bfseries{ to } $M$}
\FOR{$i=1$ \bfseries{ to } $d$}
\STATE $\mathbf{h}^i = sigm(\mathbf{v}^i\mathbf{ w}+ \mathbf{b}_h)$
\ENDFOR
\STATE $\mathbf{h}^0 = sigm(\mathbf{b}_h+\sum_{j=1}^d \mathbf{w}_j \mathbf{h}^j)$
$\hat{\mathbf{v}}^0 = \mathbf{h}^0 \mathbf{w}^\top + \mathbf{b}_v$\\
$\textrm{Error}(\mathbf{v}^0,\hat{\mathbf{v}}^0) = |\hat{\mathbf{v}}^0-\mathbf{v}^0|^2$\\
$\Delta \mathbf{w}_d  = \eta \, \partial \textrm{Error}/\partial \mathbf{w}_d$
\ENDFOR
\ENDFOR
\end{algorithmic}

\end{algorithm}
One can regard the weights $w$ as a representation of the static patterns contained in the data and the $w_d$ as representing the transformation undergone by these patterns over time in the data sequences. This allows us to separate the representation of form and motion in the case of natural image sequences, a desirable property that is frequently studied in natural movies (see \cite{cadieu2012}).

\section{Experiments}
\begin{figure} \begin{center}
\includegraphics[width=\columnwidth]{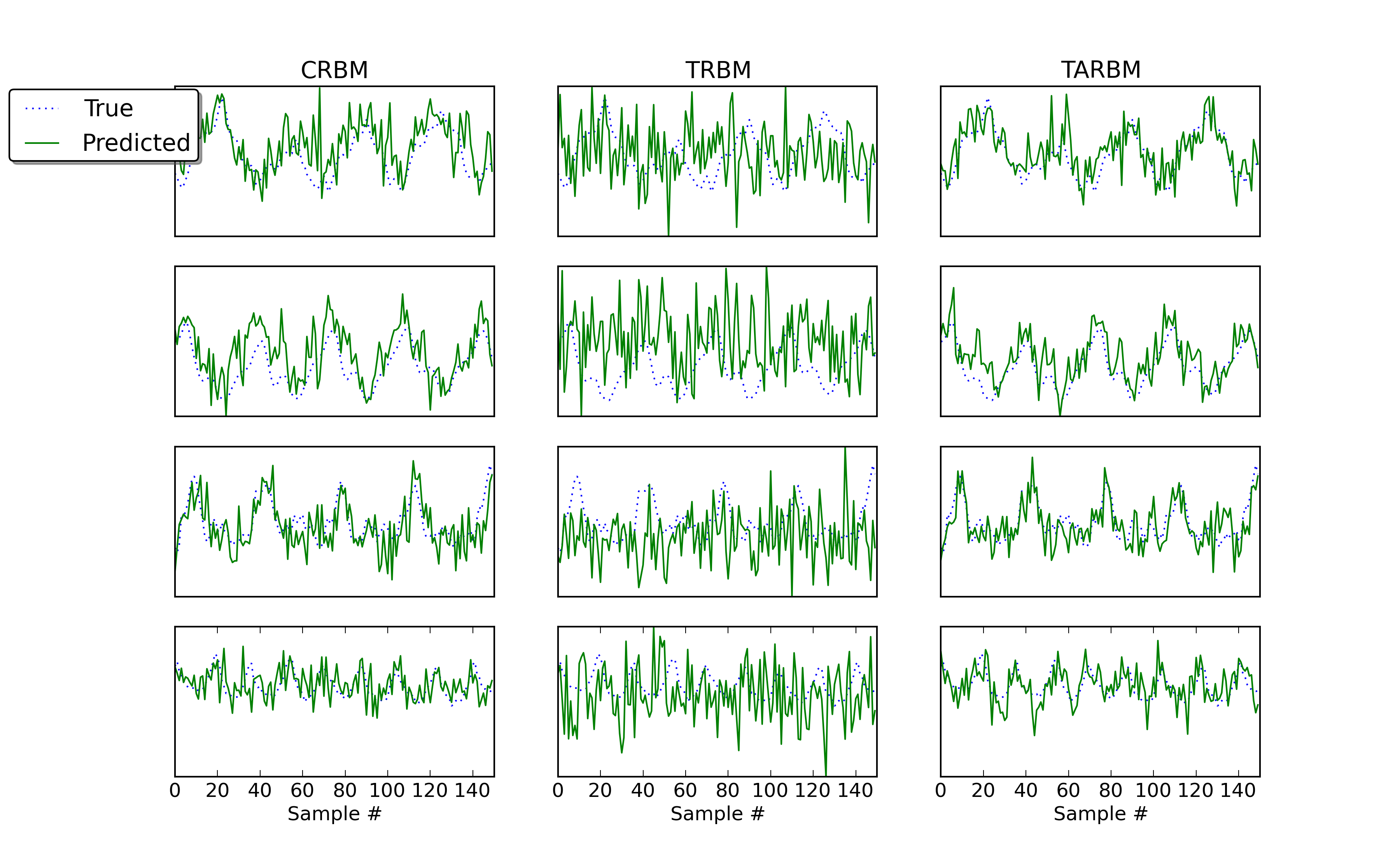}
\caption{CRBM, TRBM and TARBM used to fill in data points from motion capture data \cite{taylor2007modeling}. 4 dimensions of the motion data are shown along with the their model reconstructions from a single trial.} \label{fig:motion} \end{center}
\end{figure}

We first assess the TARBM's ability to learn multi-dimensional temporal sequences by applying it to the 49 dimensional motion capture data described in \cite{taylor2007modeling} and comparing the performance to a TRBM \footnote{In this section we refer to the reduced TRBM model referenced in \cite{sutskever2008recurrent} with only hidden to hidden temporal connections}  and Graham Taylor's example CRBM implementation \footnote{CRBM implementation available at \textit{https://gist.github.com/2505670}}. All three models are implemented using Theano \cite{theano}, have a temporal dependancy of 6 frames and were trained using minibatches of 100 samples for 500 epochs\footnote{For the TRBM, training epochs were broken up into 100 static pretraining and 400 epochs for all the temporal weights together}  \footnote{For the TARBM, training epochs were broken up into 100 static pretraining, 50 Autoencoding epochs per delay and 100 epochs for all the temporal weights together}. The training time for the models was approximately equal. Training was performed on the first 2000 samples of the dataset after which the models were presented with 1000 snippets of the data not included in training set and required to generate the next frame in the sequence. The results of a single trial prediction for 4 dimensions of the dataset can be seen in Figure \ref{fig:motion} and the mean squared error of the model predictions over 100 repetitions of the task can be seen in Table \ref{tbl:motion}. The TARBM by far outperforms the TRBM model in this task and is also somewhat better than the CRBM \footnote{No attempt was made to tune the CRBM beyond the code provided, as such it is possible that better performance could be achieved.}. The gain in performance from the TRBM to TARBM model, which are both structuraly identical, would suggest that our approach of Autoencoding the temporal dependancies gives the model a more meaningful temporal representation than is achievable through contrastive divergence alone. 
\par

\begin{table}[t]
\caption{Prediction results on the motion capture dataset}
\label{tbl:motion}
\begin{center}
\begin{tabular}{lll}
\multicolumn{1}{c}{\bf Model} &\multicolumn{1}{c}{\bf Architecture and Training} &\multicolumn{1}{c}{\bf Mean Squared Error}
\\ \hline \\
TRBM & 100 hidden units, 6 frame delay & 1.82 \\
CRBM & 100 hidden units, 6 frame delay & 0.64 \\
TARBM & 100 hidden units, 6 frame delay & 0.37 \\
\end{tabular}
\end{center}
\end{table}

The second experiment was to model a natural movie dataset and investigate the types of filters learned. Here we take the Hollywood2 dataset introduced in \cite{marszalek09}, consisting of a number of snippets from various Hollywood films and compare the CRBM implementation referenced with our TARBM model. From the dataset, 8x8 pixel patches are extracted in sequences 30 frames long. They are then contrast normalised and whitened to provide a training set of approximately 250,000 samples. The models, each with 400 hidden units and a temporal dependancy of 3 frames, are trained initially for 100 epochs on static frames of the data to initialise the $\mathbf{w}$ weights and then until convergence on the full temporal sequences.  \par

Visualisation of the temporal receptive fields learnt by the CRBM involves displaying the weight matrix $\mathbf{w}$ and the temporal weights $\mathbf{w}_{1}$ to $\mathbf{w}_{d}$ for each hidden unit as a projection into the visible layer (an 8x8 patch). This shows the temporal dependance of each hidden unit on the past visible layer activations and is plotted with time running from left to right. The visualisation process for the  TARBM is somewhat more complicated as each hidden unit is also dependant on a number of hidden units from each delay time in the model and as such cannot be visualised as a direct projection of the weights into visible layer. To understand how these units depend on the past we use a forward projection method through the temporal delays whereby a hidden unit $h$ at delay time $t-d$ is chosen as the starting point. We then use the relative weights for unit $h$ in $w_1$ to find the $n$ most likely units to be active at time $t-(d-1)$ given that unit $h$ was active at $t-d$. For each of the $n$ active units at $t-(d-1)$, we choose $n$ active units at time $t-(d-2)$ given the activations of unit $h$ at $t-d$ propogated through $\mathbf{w}_{2}$ and one of the $n$ units at $t-(d-1)$ propogated through $\mathbf{w}_{1}$. This process is repeated until the full delay of the network is mapped out. For each of the active hidden units, the projection onto an 8x8 patch of the hidden layer is defined in the weight matrix $\mathbf{w}$. When plotted for $n=1$, this trace displays the most likely evolution of the hidden layer over the delay period of the model for each hidden unit.

\begin{figure} \begin{center}
\includegraphics[width=\columnwidth]{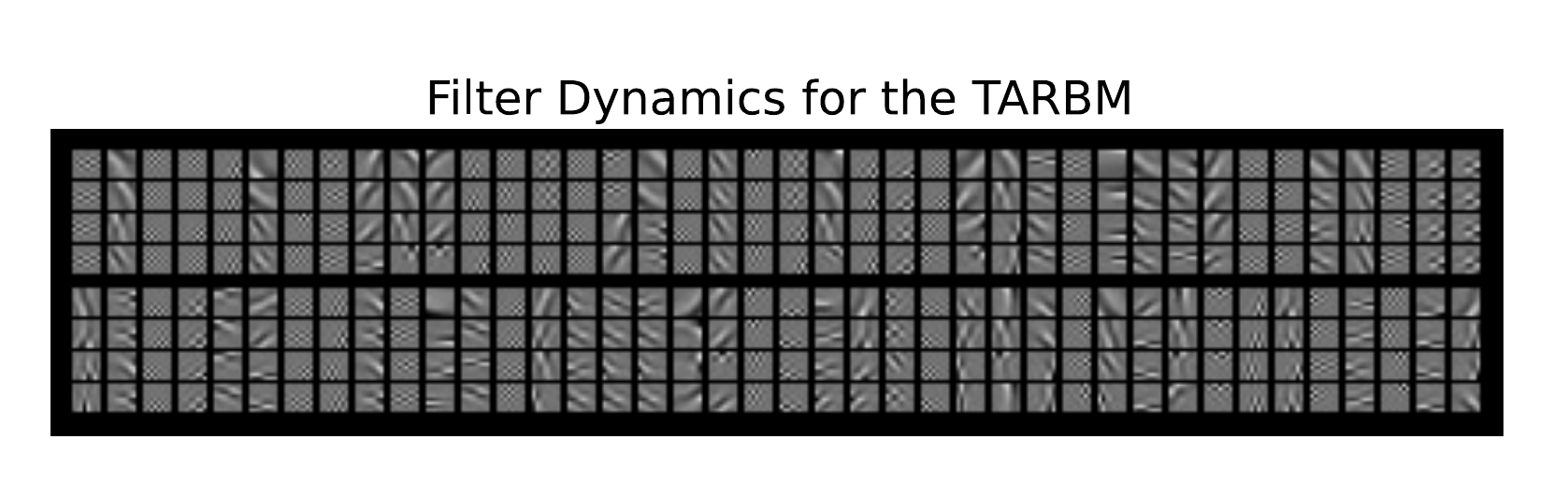}
\includegraphics[width=.99\columnwidth]{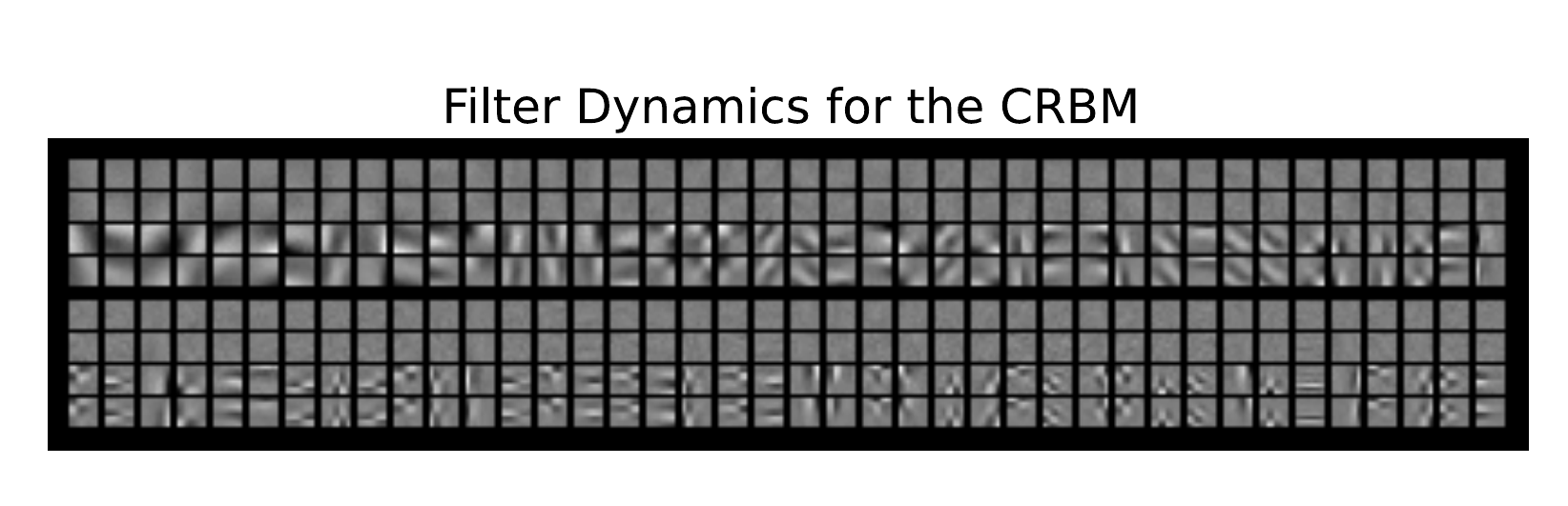}
\caption{The Temporal \emph{features} of a subset of hidden units from a TARBM  (left) and a CRBM (right). For the TARBM, we plot the most active units as described in the text ($n=1$). Each group of 4 images represents the temporal filter of one hidden unit with the lowest patch representing time $t$ and the 3 patches above representing each of the delay steps in the model. Temporal filters for the 80 units (out of 400) with highest temporal variation of the receptive fields for both models are shown. The units are displayed in two rows of 40 columns with 4 filters, with the temporal axis going from top to bottom.} \label{fig:temporal} \end{center}
\end{figure}

A subset of the temporal filters learned by each of the models can be seen in Figure \ref{fig:temporal} with the TARBM on the left and the CRBM on the right. While both the TARBM and the CRBM learn gabor like filters at time $t$, their dependance on the past is markedly different. Most hidden units in the CRBM fail to capture any structured dependance on delay times greater than $d=1$. This makes the CRBMs temporal filters difficult to interperet with respect to structure in the image. The layerwise training of the temporal weights in the TARBM along with the forced reliance on filters learned in $w$ for its delay input give the TARBM not only a longer temporal dependance, but also allow the weights learned to be easily interpereted as a transformation of the learned filters. 

\begin{figure} \begin{center}
\includegraphics[width=\columnwidth]{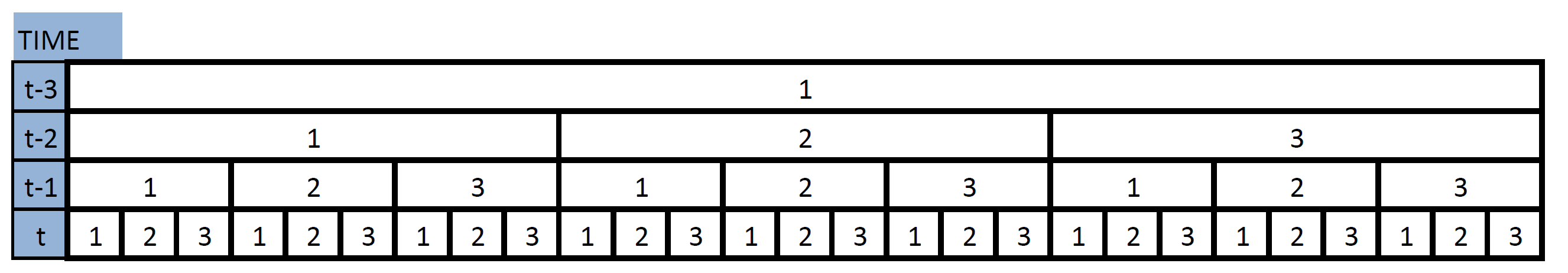}
\includegraphics[width=\columnwidth]{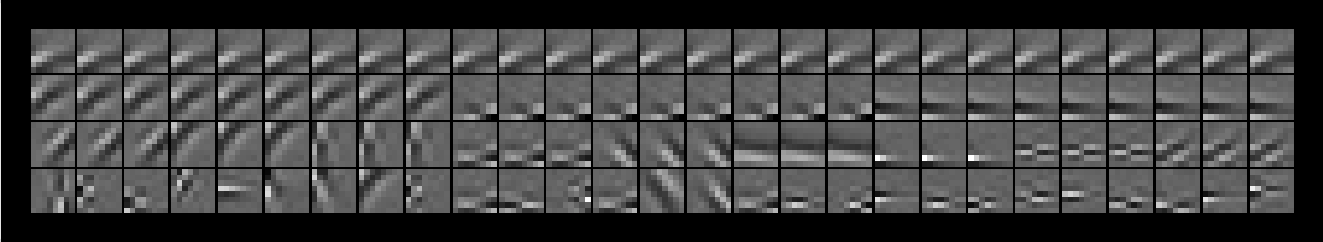}
\includegraphics[width=\columnwidth]{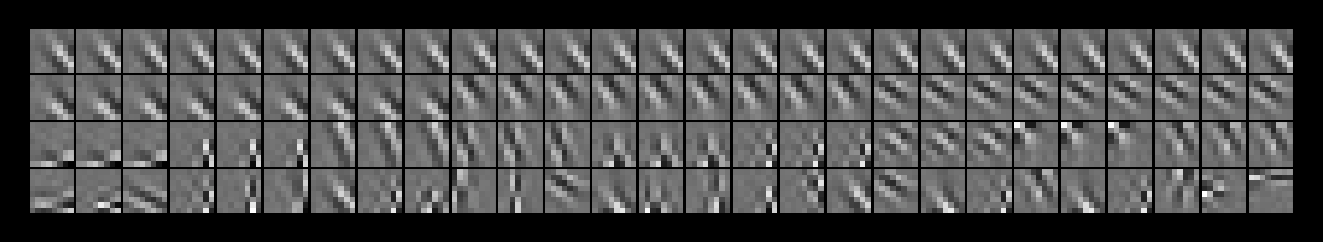}
\includegraphics[width=\columnwidth]{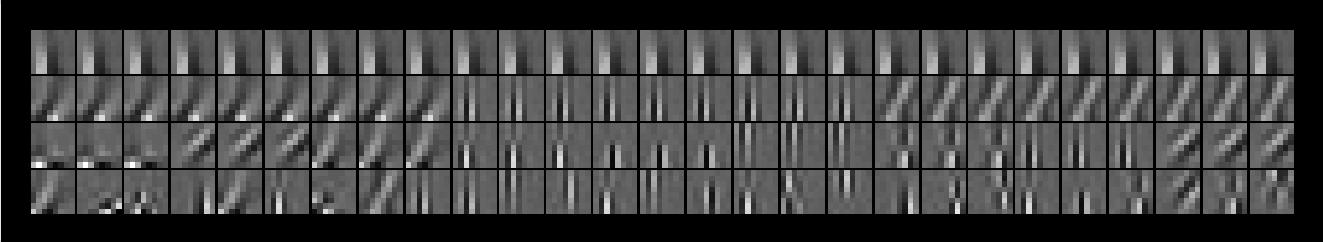}
\caption{Temporal Filters of 3 hidden units in the TARBM after training on the Hollywood2 dataset ($n=3$). The top image shows the schematic of the three images below it. Each patch in the top row of an image represents the activation of a single hidden unit at time $t-3$ where $d=3$ is the delay of the TARBM. The second row down shows the 3 most likely units to be activated at $t-2$ given the activation of the unit at $t-3$ and so on for the 3rd and 4th rows, forming a tree structure of dependancy. For ease of interpritation, units with multiple descendants are repeated so that each column can easily be read top to bottom.} \label{fig:temporal_joined} \end{center}
\end{figure}

Figure \ref{fig:temporal_joined} shows the forward projection method of visualising the TARBM for $n=3$ from selected hidden units. This means that for each delay step, three most likely filters to be active at the next point in time are shown. The model is able to learn multiple trainformations over time for each of the hidden unit receptive fields. The transformations often represent simple operations such as rotation and translation of the static features, seperating the modeling of form and motion.
\par

\section{Discussion and Future Work}

We have shown that by using an Autoencoder to initialise the temporal weights of a TRBM, forming what we call a TARBM, a significant performance increase can be achieved in modelling and generating from a sequential motion capture dataset. We also show that the TARBM is able to learn high level structure from natural movies and account for the transformation of these features over time. Additionally, the evolution of the learned temporal filters are easily interpretable and help to better understand how the model represents the trained data.\par
The presented model could with minimal effort be adapted into a deep architecture, allowing us to represent higher order features in the same temporal manner. We propose that learning higher order temporal features might prove to be useful for control tasks such as image stabilization and object tracking. In addition, we hope to study the relation of the presented encoding strategy with strategies employed by the mammalian visual cortex \cite{kampa2011}. Another interesting avenue of research will be to apply the current model to classification and generative tasks.

\subsubsection*{Acknowledgments}

The work of Chris H\"ausler and Alex Susemihl were supported by the DFG Research Training Group 1589/1. We would like to thank Prof. Martin Nawrot of the Freie Universit\"at Berlin for his enthusiastic support and invaluable feedback.

\bibliography{library}{}
\bibliographystyle{unsrt}

\end{document}